\documentclass[conference]{IEEEtran}

\AtEndDocument{\par\leavevmode}
\IEEEoverridecommandlockouts
\usepackage{cite}
\usepackage{amsmath,amssymb,amsfonts,amsthm}
\usepackage{graphicx}
\usepackage{textcomp}
\usepackage{xcolor}

\usepackage{flushend}
\usepackage{times, soul, url}
\usepackage[utf8]{inputenc}
\usepackage[small]{caption}
\usepackage{booktabs, wrapfig}
\usepackage{algorithm, algpseudocode}
\usepackage{tikz}
\usetikzlibrary{shapes.multipart}
\usetikzlibrary{decorations.pathreplacing}
\usepackage{textcmds, nicefrac}
\usepackage{hang, enumitem}
\algrenewcommand{\algorithmicrequire}{\textbf{Input:}}
\algrenewcommand{\algorithmicensure}{\textbf{Output:}}
\algrenewcommand{\algorithmiccomment}[1]{//#1}
\algrenewcommand\algorithmicindent{0.7em}

\usepackage{array}
\newcolumntype{N}{>{\centering\arraybackslash}m{2.37cm}}
\newcolumntype{C}{>{\centering\arraybackslash}m{1.7cm}}
\newcolumntype{L}{>{\centering\arraybackslash}m{0.8cm}}
\usepackage{multirow}

\def\BibTeX{{\rm B\kern-.05em{\sc i\kern-.025em b}\kern-.08em
    T\kern-.1667em\lower.7ex\hbox{E}\kern-.125emX}}

\begin{document}
\title{Evolutionary NAS with Gene Expression Programming of Cellular Encoding
\thanks{This work was supported in part by the Soka Makiguchi  Foundation for Education, Japan.}
}

\author{\IEEEauthorblockN{Cliford Broni-Bediako}
\IEEEauthorblockA{\textit{Graduate School of Science and Engineering} \\
	\textit{Soka University, Hachioji, Tokyo, Japan}\\
	e18d5252@soka-u.jp} \\
\IEEEauthorblockN{Luiz H. B. Mormille}
\IEEEauthorblockA{\textit{Graduate School of Science and Engineering} \\
	\textit{Soka University, Hachioji, Tokyo, Japan}\\
	e18d5251@soka-u.jp} 
\and
\IEEEauthorblockN{Yuki Murata}
\IEEEauthorblockA{\textit{ Graduate School of Science and Engineering} \\
	\textit{Soka University, Hachioji, Tokyo, Japan}\\
	e19d5202@soka-u.jp} \\
\IEEEauthorblockN{Masayasu Atsumi}
\IEEEauthorblockA{\textit{Graduate School of Science and Engineering} \\
	\textit{Soka University, Hachioji, Tokyo, Japan}\\
	matsumi@soka.ac.jp}}

\maketitle
\IEEEpubidadjcol
% change the default value for CTLdash_repeated_names
\bstctlcite{IEEEexample:BSTcontrol}

\begin{abstract}
The renaissance of neural architecture search (NAS) has seen classical methods such as genetic algorithms (GA) and genetic programming (GP) being exploited for convolutional neural network (CNN) architectures. While recent work have achieved promising performance on visual perception tasks, the direct encoding scheme of both GA and GP has functional complexity deficiency and does not scale well on large architectures like CNN. To address this, we present a new generative encoding scheme\textemdash{\it symbolic linear generative encoding} (SLGE)\textemdash simple, yet a powerful scheme which embeds local graph transformations in chromosomes of linear fixed-length string to develop CNN architectures of variant shapes and sizes via an evolutionary process of gene expression programming. In experiments, the effectiveness of SLGE is shown in discovering architectures that improve the performance of the state-of-the-art handcrafted CNN architectures on CIFAR-10 and CIFAR-100 image classification tasks; and achieves a competitive classification error rate with the existing NAS methods using fewer GPU resources.
\end{abstract}

\begin{IEEEkeywords}
neural architecture search; convolutional neural network; gene expression programming; cellular encoding 
\end{IEEEkeywords}

\section{Introduction}
Historically, evolutionary neural architecture search (NAS) research has been of interest in the AI community for three decades~\cite{yao_1999:e-ann}. Recent growing interest in automation of deep learning has seen a phenomenal development of new NAS methods based on reinforcement learning (RL) and others in the last three years~\cite{elsken_2019a:nas-s}. However, classical NAS methods such as Genetic Algorithms (GA)~\cite{xie_2017:g-cnn,sun_2019a:ca-cnn} and Genetic Programming (GP)~\cite{suganuma_2018:gp-cnna} has also been exploited to develop convolutional neural networks (CNN) for visual perception tasks. Besides achieving promising results, classical NAS methods also consume fewer computational resources~\cite{sun_2019a:ca-cnn} than their RL-based competitors~\cite{zoph_2017:nas-rl}. But, the direct encoding scheme of both GA and GP has two inherent limitations: (1) chromosomes of fixed-length and (2) genotype and phenotype spaces not distinctive separated, which limit their functional complexity~\cite{ferreira_2006:gep}. For example, chromosomes of fixed-length scheme employed in Genetic CNN~\cite{xie_2017:g-cnn} does not scale well on large architectures. 

To address these issues, EvoCNN~\cite{sunEvoCNN:2020} introduced variable-length scheme in GA, and was adopted in CA-CNN~\cite{sun_2019a:ca-cnn}. The variable-length scheme is used to encode CNN architectures of variant shapes and sizes similar to nonlinear structures in GP. Though, the variable-length scheme exhibits some certain amount of functional complexity, it is not without difficulty to reproduce with crossover operations. The reason for the crossover limitation in both EvoCNN and CA-CNN, and their alike CGP-CNN~\cite{suganuma_2018:gp-cnna} is because their genotype and phenotype spaces are not explicitly separated, thus, genetic modification is directly subject to phenotype structural constraint. DENSER~\cite{assuno_2019:denser} combines GA with Grammatical Evolution (GE) to separate  genotype and phenotype spaces in analogous to nature. However, GE lacks modularity and it is not flexible to modify the grammar with modules~\cite{swafford_2011:e-ge}. To this end, we argue that the development of evolutionary NAS methods that separate genotype and phenotype spaces to generate CNN architectures for visual perception tasks is still in its infancy. 

In this work, we introduce a new generative encoding scheme, {\it symbolic linear generative encoding}, which embeds local graph transformations of Cellular Encoding~\cite{gruau_1994:thesis} in simple linear fixed-length chromosomes of Gene Expression Programming~\cite{ferreira_2006:gep} to develop CNN architectures of variant shapes and sizes. Moreover, to enable the evolutionary process to discover new motifs for the architectures, {\it regular convolution}\footnote{We define a regular convolution operation as a standard convolution operation with batch normalization and ReLU.} operations are employed as the basic search units as opposed to sophisticated building-blocks such as ResNet and DenseNet blocks in CA-CNN~\cite{sun_2019a:ca-cnn}. 

In experiments, the preliminary results show the effectiveness of the proposed method by discovering CNN architecture that obtains 3.74\% error on CIFAR-10 image dataset benchmark and 22.95\% error when transfer to CIFAR-100. The results is competitive with the current auto-generated CNN architectures and an improvement to the performance of the state-of-the-art handcrafted ones. The remainder of the paper is organised as follows. Section 2 discusses the related work and Section 3 presents the details of the proposed method. The experimental results are reported in Section 4 and Section 5 concludes this study with future direction.		

\section{Related Work}
\subsection{Evolutionary NAS}
Most earlier work in evolutionary NAS evolve both the network architecture and its connection weights at a small scale~\cite{yao_1999:e-ann}. Recent neural networks such as CNN have been scaled up with millions of connection weights to improve performance on a given task. And learning the connection weights of these large-scale networks via back-propagation method outperform evolutionary approach. Thus, recent evolutionary NAS work~\cite{elsken_2019a:nas-s} have focused on evolving only the network architectures and using back-propagation method to optimize the connection weights. 

Typically, chromosomes of fixed-length is used to represent the network architectures~\cite{desell_2017:ls-ecnn,xie_2017:g-cnn,real_2019:re-icas}. But, since the best architecture shape and size for a given data is not known, chromosomes of variable-length using direct encoding scheme have been employed for the architectures to adapt their shapes and sizes on a given task ~\cite{real_2017:l-eic,liu_2018:hr-eas,sun_2019a:ca-cnn}. CGP-CNN~\cite{suganuma_2018:gp-cnna} used Cartesian GP to represent CNN architectures of variable-length structures. To depart from the functional complexity deficiency of direct encoding, DENSER~\cite{assuno_2019:denser} combines GA with GE to adopt a genotype and phenotype spaces distinction and explicitly use grammars to generate phenotypes of CNN architectures. 

Generally, the architecture search space is classified into two categories: the global search space which defines the entire architecture (macroarchitecture)~\cite{real_2017:l-eic,sun_2019a:ca-cnn}, and cell-based search space for discovering a microarchitecture (cell) which can be stacked repeatedly to build the entire architecture~\cite{liu_2018:hr-eas,real_2019:re-icas}. The cell is a directed acyclic graphs (DAGs) which is used as building blocks to form the architecture. CNN architectures discovered by cell-based approach are flexible and transferable to other tasks~\cite{zoph_2018:lta}, and they perform better than the global ones~\cite{pham_2018:e-nas-ps}. The popular cell-based approach is NASNet~\cite{zoph_2018:lta} which involves two types of cells: the normal cell and reduction cell (used to reduce feature resolutions). Zhong {\it et al.}~\cite{zhong_2018:p-nnag} and Liu {\it et al.}~\cite{liu_2018:hr-eas} proposed similar cell-based approach but used max-pooling and separable convolution layers respectively to reduce the feature resolutions. The basic search units in both search spaces are mostly sophisticated convolutions such as depthwise separable and asymmetric convolutions in AmoebaNet~\cite{real_2019:re-icas} or sophisticated blocks such as ResNet and DenseNet in CA-CNN~\cite{sun_2019a:ca-cnn}. These search units reduce the search space complexity, however, they may detriment the flexibility and restrict the discovering of new architecture building blocks that can improve the current handcrafted ones. Thus, in this work, we use regular convolutions (Section \ref{sec:search-space}).
\subsection{Gene Expression Programming}
Gene Expression Programming (GEP) is a full-fledged genotype-phenotype evolutionary method based on both GA and GP. Chromosomes consist of linear fixed-length genes similar to the ones used in GA, and are developed in phenotype space as expression-trees of different shapes and sizes similar to the parse-trees in GP. The genes are structurally organized in a head and a tail format called Karva notation~\cite{ferreira_2006:gep}. The separation of genotype and phenotype spaces with distinct functions allows GEP to perform with high effectiveness and efficiency that surpasses GA and GP~\cite{zhong_2017:gep-s}.

Ferreira~\cite{ferreira_2006:gep} has proposed that the chromosomes in GEP can completely encode ANN to discover an architecture via evolutionary process. Thus far, GEP is yet to be adopted for complex architectures like CNN. The Achilles’ heel of GEP is its Karva notation which does not allow hierarchical composition of candidate solutions, which means an evolved good motifs are mostly destroyed by genetic modifications in subsequent generations~\cite{li_2005:p-gep}. Thus, to adopt GEP for CNN architecture search, we propose a new generative scheme that naturally embeds motifs in Karva expression of GEP as individual chromosomes and consequently a new genotype-phenotype mapping is encapsulated in GEP convention of evolutionary process. 
\subsection{Cellular Encoding}
Cellular Encoding (CE) is a generative encoding based on simple local graph transformations that control the division of nodes which evolve into ANN of different shapes and sizes. The graph transformations are represented by program symbols with unique names; and these symbols depict a grammar-tree (program) which encapsulates the developing procedure of an ANN via evolutionary process from a single initial unit\footnote{A unit is a single neuron in the original idea of CE, but in this work, it represents a convolutional layer of neurons} that has both input and output nodes. CE has shown its efficiency on a wide range of problems such as evolving ANN for controlling two poles on a cart and locomotion of a 6-legged robot, a review can be found in Gruau~\cite{gruau_1996c:a-cd}. 

\begin{figure}[h]
	\centering
	\begin{tikzpicture}[line width=0.5pt, yscale=.65pt]
		\tikzstyle{io}=[draw, rounded corners, text width=0.9cm, text centered, text height=5pt, fill=gray!30]
		\tikzstyle{pc}=[draw, text width=1cm, text centered, text height=5pt]
		%\draw[style=help lines] (0,0) grid(6, 3);
		
		%% SQE illustration
		\node[io] (in) at (1,3){$input$};
		\node[pc] (p) at (1,2){parent};
		\node[pc] (c) at (1,1){child};
		\node[io] (out) at (1,0){$output$};
		\draw[->] (in) edge (p)
		(p) edge (c)
		(c) edge (out);
		
		%%CPI illustration
		\node[io] (in) at (3,3){$input$};
		\node[pc] (p) at (3,2){parent};
		\node[pc] (c) at (3,1){child};
		\node[io] (out) at (3,0){$output$};
		\draw[->] (in) edge (p)
		edge[bend right=68] (c)
		(p) edge (c)
		(c) edge (out);
		
		%%CPO illustration
		\node[io] (in) at (5,3){$input$};
		\node[pc] (p) at (5,2){parent};
		\node[pc] (c) at (5,1){child};
		\node[io] (out) at (5,0){$output$};
		\draw[->] (in) edge (p)
		(p) edge (c)
		edge[bend right=68] (out)
		(c) edge (out);

		\draw[decorate] (1,3.3) node[above] {SQE};
		\draw[decorate] (3,3.3) node[above] {CPI};
		\draw[decorate] (5,3.3) node[above] {CPO};
	\end{tikzpicture}
	\caption{Illustration of SEQ, CPI and CPO transformation functions.}
	\label{fig:ce-func}
\end{figure}
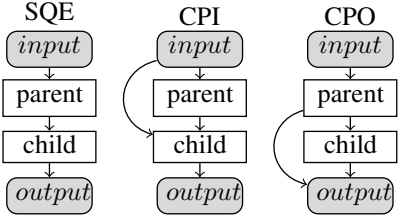

In this work, we adopt four transformation functions of CE and embed them in the fixed-length individual chromosomes of GEP, since their transformations\textemdash particularly $CPI$ and $CPO$~\cite{Gruau:1996:CEI:869639}\textemdash generate motifs similar to ResNet block which is prominent in deep neural networks. We briefly explain the four functions we adopt and refer Gruau~\cite{gruau_1994:thesis} and Gruau and Quatramaran~\cite{Gruau:1996:CEI:869639} for in-depth details. Fig. \ref{fig:ce-func} depicts the graphical representation of the SEQ, CPO and CPI transformation functions.
\begin{itemize}[noitemsep,topsep=0pt,parsep=0pt,partopsep=0pt]
	\item[--]SEQuential division $(SEQ)$: it splits current node into two and connects them in serial; the child node inherits the outputs of the parent node.
	\item[--]CoPy Input division $(CPI)$: it performs $SEQ$, then shares the same inputs with parent and child nodes.
	\item[--]CoPy Output division $(CPO)$: it performs $SEQ$, then shares the same outputs with parent and child nodes.
	\item[--]END program $(END)$: it stops the developing process.
\end{itemize}

\section{Proposed Method}
We formulate the CNN architecture search problem as follows. Given the problem space $\Psi=\{\mathcal{A, S, P}, t\mathcal{D}, v\mathcal{D}\}$, where $\mathcal{A}$ is the architecture search space, $\mathcal{S}$ represents the search strategy, $\mathcal{P}$ denotes the performance measure, and $t\mathcal{D}$ and $v\mathcal{D}$ are the training and validation datasets respectively, the objective is to find a smaller CNN architecture $a^*\in \mathcal{A}$ via the search strategy $\mathcal{S}$, then after training it on the dataset $t\mathcal{D}$, it maximizes some performance $\mathcal{P}$ (in this case classification accuracy $\mathcal{P}_{acc}$) on the validation dataset $v\mathcal{D}$. The smaller architecture here means a model with fewer number of parameters $\theta$. Mathematically, the objective function $\mathcal{F}$ can be formulated as:
\begin{align}
\begin{aligned}
\mathcal{F}(\Psi) & = \underset{\theta, a}{max}\ \ {\mathcal{P}_{acc}(\mathcal{L}(a(\theta), t\mathcal{D} \ |\ \mathcal{S}, a\in \mathcal{A}), v\mathcal{D})} \\
s.t. \ \ & ModelSize(a(\theta)) \leq\mathcal{T}_{params}
\end{aligned}
\label{eq:objfunc}
\end{align}
where $\mathcal{L}$ represents the training of the model parameters $\theta$ with the loss function and $\mathcal{T}_{params}$ denotes the target number of parameters. In this section, we describe the architecture search space and search strategy that we proposed.
\subsection{Search Space}\label{sec:search-space}
The basic search units in the search space consists of regular convolutions with batch-norm and ReLU and CE program symbols. The regular convolutions units may enable the evolutionary process to find new motifs to form CNN architectures rather than the predefined ones (depthwise separable, asymmetric convolution, ResNet and DenseNet). Since predefined units reduce the complexity of the architecture search space, they may detriment the flexibility of discovering new motifs that can improve the current handcrafted ones. Thus, we adopt cell-based search approach~\cite{zoph_2018:lta} in which each node in the discovered cells is associated with a regular convolution in the search space whereas edges are representation of latent information flow direction. Following Zhong {\it et al.}~\cite{zhong_2018:p-nnag}, we use max-pooling to reduce feature maps resolution and 1$\times$1 convolution is applied when necessary to downsample the input depth. 

The regular convolutions include in the search space are: 1$\times$1, 1$\times$3, 3$\times$1, and 3$\times$3. Each operation has a stride of one and appropriate pad is applied to preserve the spatial resolution of the feature maps. And the CE program symbols are: $SEQ$, $CPI$, $CPO$ and $END$. The complexity of the search space can be expressed as: $(\#porgram\_symbols)^h\times (\#convolution\_operations)^{h+1}\times n$ possible architectures, where $h$ is the number of CE program symbols that forms the head of a gene in a chromosome and $n$ is the number of genes in a chromosome. For example, a chromosomes with $h$=2 and $n$=3, the search space contains 3072 possible architectures.
\subsection{Symbolic Linear Generative Encoding} 
Encoding network architectures into genotypes using a particular encoding scheme is the first stage of evolutionary NAS task. We proposed a new generative encoding scheme, {\it symbolic linear generative encoding} (SLGE), which embeds local graph transformations of CE in simple linear fixed-length chromosomes of GEP to develop CNN architectures of different shapes and sizes. Chromosomes in SLGE can evolve different motifs via the evolutionary process of GEP to build CNN architectures. SLGE explicitly separates genotype and phenotype spaces in analogue to nature, to benefit from all the advantages of evolutionary process without functional complexity deficiency. It is worth emphasizing how the implementation of SLGE is simple, because of its linear fixed-length structure. We implement SLGE on the top of geppy\footnote{https://geppy.readthedocs.io/en/latest/index.html}, a GEP library based on DEAP\footnote{https://deap.readthedocs.io/en/master/} framework.
\subsubsection{Representation}
Genetic representation defines the encoding of a phenotype into a genotype. A simple, but effective and efficient representation significantly affects the overall performance of evolutionary process. In SLGE, the chromosomes are structured similar to the ones in GEP. Chromosomes are made up of genes of equal fixed-length string. Each gene composes of a head which consists of CE program symbols, and a tail of regular convolutions. Given the length of a gene head $h$, the length of the gene tail $t$ is a function of $h$ expressed as: $t=h+1$, thus, the length of a gene is $2h +1$. Fig. \ref{fig:chrom} is an example of a typical SLGE chromosome of two genes and its phenotype (cell) is presented in Fig. \ref{fig:cell}. 
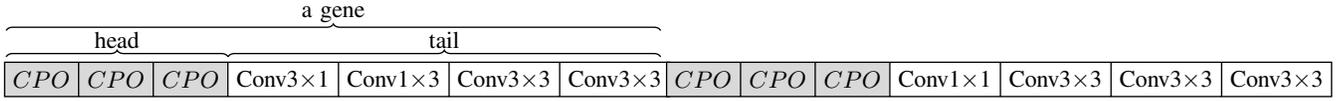
\begin{figure*}
	\centering
	\fontsize{9}{0}
	\begin{tikzpicture}[line width=0.5pt]
	% gene 0
	%\draw[style=help lines] (-4,0) grid(13,1);
	\node [rectangle split, rectangle split parts=7, rectangle split horizontal, 
	rectangle split part fill={gray!30, gray!30, gray!30, white}, draw]
	at (0,0) {$CPO$\nodepart{two}$CPO$\nodepart{three}$CPO$\nodepart{four}Conv3$\times$1
		\nodepart{five}Conv1$\times$3\nodepart{six}Conv3$\times$3\nodepart{seven}Conv3$\times$3};
	%		\node [rectangle split, rectangle split parts=4, rectangle split horizontal, draw, densely dotted]
	%		at (0,-1) {Conv3x1\nodepart{two}Conv1x1\nodepart{three}Conv3x3\nodepart{four}Conv1x1};
	%		\draw[densely dotted] (0.16,-0.25) -- (-2.75,-0.77); 
	%		\draw[densely dotted] (2.74,-0.25) -- (2.74,-0.77);
	\draw[decorate, decoration=brace] (-4.4,0.65) -- node[above] {a gene} (4.3,0.65);
	\draw[decorate, decoration=brace] (-4.4,0.3) -- node[above] {head}  (-1.45,0.3);
	\draw[decorate, decoration=brace] (-1.45,0.3) -- node[above] {tail}  (4.3,0.3);
	% gene 1
	\node [rectangle split, rectangle split parts=7, rectangle split horizontal, 
	rectangle split part fill={gray!30, gray!30, gray!30, white}, draw]
	at (8.8,0) {$CPO$\nodepart{two}$CPO$\nodepart{three}$CPO$\nodepart{four}Conv1$\times$1
		\nodepart{five}Conv3$\times$3\nodepart{six}Conv3$\times$3\nodepart{seven}Conv3$\times$3};
	%		\node [rectangle split, rectangle split parts=4, rectangle split horizontal, draw, densely dotted]
	%		at (5.5,1) {Conv1x1\nodepart{two}Conv3x3\nodepart{three}Conv3x3			\nodepart{four}Conv3x1};
	%		\draw[densely dotted] (5.61,0.25) -- (2.75,0.77); 
	%		\draw[densely dotted] (8.1,0.25) -- (8.24,0.77);
	% gene 2		
	%		\node [rectangle split, rectangle split parts=7, rectangle split horizontal, 
	%		rectangle split part fill={gray!30, gray!30, gray!30, white}, draw]
	%		at (10.86,0) {$CPO$\nodepart{two}$SEQ$\nodepart{three}$END$\nodepart{four}$C0$
	%			\nodepart{five}$C1$\nodepart{six}$C2$\nodepart{seven}$C3$};
	%		\node [rectangle split, rectangle split parts=4, rectangle split horizontal, draw, densely dotted]
	%		at (11,-1) {Conv1x1\nodepart{two}Conv1x3\nodepart{three}Conv3x1
	%			\nodepart{four}Conv1x1};
	%		\draw[densely dotted] (11.0,-0.25) -- (8.25,-0.78); 
	%		\draw[densely dotted] (13.65,-0.25) -- (13.74,-0.77);
	\end{tikzpicture}
	\caption{The structural representation of a typical SLGE chromosome. The chromosome has two genes of equal fixed-length, and each gene has a head of three CE program symbols and a tail of four regular convolution operations. This chromosome is the genotype of the cell (Fig. \ref{fig:cell}) used to build the best evolutionary-discovered network in Table \ref{tab:result1}.}
	\label{fig:chrom}
\end{figure*}
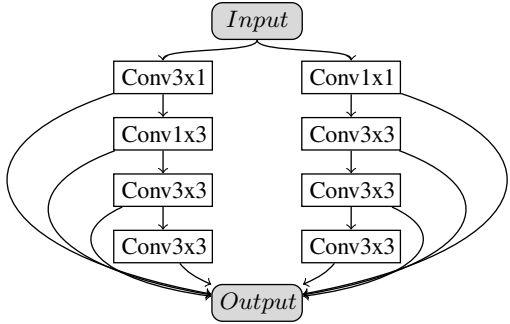
\begin{figure}
	\centering
	\fontsize{9}{0} %xscale=1.1pt, yscale=.021cm,  line width=0.5pt
	\begin{tikzpicture}[line width=0.5pt, xscale=2.5pt, yscale=.75pt]
	\tikzstyle{inout}=[draw, rounded corners, text width=0.99cm, text centered, fill=gray!30]
	\tikzstyle{conv}=[draw, text width=1.1cm, text centered]
	%\draw[style=help lines] (0,0) grid(1,5);
	\node[inout] (0) at (0.5,5) {$Input$};
	\node[inout] (10) at (0.5,0) {$Output$};
	% gene 0
	\node[conv] (1) at (0,4) {Conv3x1};
	\node[conv] (2) at (0,3) {Conv1x3};
	\node[conv] (3) at (0,2) {Conv3x3};
	\node[conv] (4) at (0,1) {Conv3x3};
	\draw[->] (0) edge[out=south, in=north] (1)
	(1) edge (2)
	(1) edge[bend right=50] (10)
	(2) edge (3)
	(2) edge[bend right=50] (10)
	(3) edge (4)
	(3) edge[bend right=50] (10)
	(4) edge[bend right=10] (10);
	
	% gene 1
	\node[conv] (5) at (1,4) {Conv1x1};
	\node[conv] (6) at (1,3) {Conv3x3};
	\node[conv] (7) at (1,2) {Conv3x3};
	\node[conv] (8) at (1,1) {Conv3x3};
	\draw[->] (0) edge[out=south, in=north] (5)
	(5) edge (6)
	(5) edge[bend left=50] (10)
	(6) edge (7)
	(6) edge[bend left=50] (10)
	(7) edge (8)
	(7) edge[bend left=50] (10)
	(8) edge[bend left=10] (10);
	
	% gene 2
	%		\node[conv] (7) at (2,4) {Conv1x1};
	%		\node[conv] (8) at (2,3) {Conv1x3};
	%		\node[conv] (9) at (2,2) {Conv3x1};
	%		\draw[->] (0) edge[out=south, in=north] (7)
	%		(7) edge (8)
	%		edge[bend left=35] (9)
	%		(8) edge (9)
	%		(9) edge[out=south, in=40] (10);
	\end{tikzpicture}
	\caption{The representative cell of the chromosome in Fig. \ref{fig:chrom}, and it is the cell used to build the top network in Table \ref{tab:result1}. (In general, the convolution nodes without successor in the cell are depthwise concatenated to provide an output, and if a node has more than one predecessor, the predecessors parameters are add together. This is a common approach in cell-based search~\protect\cite{zoph_2018:lta}.) }
	\label{fig:cell}
\end{figure}
\subsubsection{Mapping and Fitness Function}
The mapping algorithm translates genotypes (chromosomes) into phenotypes (cells)\textemdash each one is stacked repeatedly to build candidate CNN architecture\textemdash then the fitness function is used to train each architecture for a few epochs and evaluates to determine its fitness quality in the phenotype space, which is then maps back into genotype space where genetic variations occur to produce offspring for the next generation. The representation can be mathematically expressed as follows:
\begin{align}
\label{eq:mapfunc}
\mathcal{F}_g(\varphi_g)&:\Phi_g\rightarrow\Phi_p \\
\mathcal{F}_p(\varphi_p)&:\Phi_p\rightarrow \mathbb{R}
\label{eq:fitfunc}
\end{align}
where $\mathcal{F}_g$ is the mapping algorithm which maps a set of genotypes $\varphi_g\subset\Phi_g$ into phenotype space $\Phi_p$ to form individual architectures and $\mathcal{F}_p$ represents the fitness function that trains each individual architecture (phenotype) $\varphi_p\in\Phi_p$ for a few epochs to determine its fitness value in the fitness space $\mathbb{R}$. Algorithm \ref{alg:mapping} and \ref{alg:ceprogram} are the mapping algorithm $\mathcal{F}_g$ that is used to translate chromosomes into cells in SLGE. The algorithm takes a chromosome of linear fixed-length string as an input and produces a DAG which represents a candidate cell. For example, given the chromosome in Fig. \ref{fig:chrom} as an input, Algorithm \ref{alg:mapping} and \ref{alg:ceprogram} will produce the cell in Fig. \ref{fig:cell} as an output.

The fitness function $\mathcal{F}_p$ is a surrogate to the objective function $\mathcal{F}$ in equation (\ref{eq:objfunc}) which aims to find an individual architecture that maximize classification accuracy on validation dataset, subject to model size constraint. However, since we search for cells in the architecture search space during the evolutionary process, the  model size constraint of individuals is not considered in their fitness evaluation. Rather, we only apply the model size constraint for the entire architecture. Thus, the function is simply the objective function $\mathcal{F}$ in equation (\ref{eq:objfunc}) without the size constraint. The loss function is used to train each individual via back-propagation on the training and validation sets, and the fitness function determines its fitness value which represents the probability of the individual survival. The higher the fitness value, the more likely the individual will have progeny and survive into the next generation.
\begin{algorithm}[tb]
	\caption{The genotype-phenotype mapping function $\mathcal{F}_g$}
	\label{alg:mapping}
	\begin{algorithmic}[1] 
		\Require Chromosome of linear fixed-length string $\varphi_g\in\Phi_g$
		\Ensure A directed acyclic graph $DAG$ (cell)
		\State $n\gets len(\varphi_g)$	\hskip1.9em\Comment{Number of genes in chromosome $\varphi_g$}
		\State $DAG.init(null)$   \hskip0.5em\Comment{Initialize cell with $null$ node which}
		\Statex \hskip7.7em\Comment{has input and output nodes}
		\For {$i\gets 1$ \textbf{to} $n$}
		\State $\varphi_g^{(i)}\gets \varphi_g[i]$ \hskip1.5em\Comment{Get gene $i$ in chromosome $\varphi_g$}
		\State Create queue $Q$ of all the convolutions in gene $\varphi_g^{(i)}$
		\State $pnode\leftarrow Q.dequeue(0)$  \hskip0.0em\Comment{Parent node}
		\State $cnode\leftarrow Q.next()$    \hskip2.1em\Comment{Child node}
		\State $G\gets subgraph(pnode)$  \hskip0.3em\Comment{Initialize subgraph of $DAG$} 
		\Statex \hskip11em\Comment{for $\varphi_g^{(i)}$ with $pnode$ which}	
		\Statex \hskip11em\Comment{has input and output nodes}	
		\State $pos\gets0$ 
		\While {$|Q|> 0$}
		\State $ps\gets \varphi_g^{(i)}[pos]$  \hskip3.2em\Comment{Get CE program symbol}
		\If {$ps = ${\it\qq{END}}}
		\State $DAG.merge(G)$ \hskip1.7em\Comment{Merge subgraph $G$ to $DAG$}
		\Statex \hskip11em\Comment{at input and output nodes}
		\State \textbf{return}
		\EndIf
		\State \Call{Transform}{$G, ps, pnode, cnode$} \hskip1em\Comment{Algorithm \ref{alg:ceprogram}}
		\State $pnode\gets Q.dequeue(0)$ 
		\State $cnode\gets Q.next()$ 
		\State $pos\gets pos+1$
		\EndWhile
		\State $DAG.merge(G)$ 
		\EndFor
		\State \textbf{return} $DAG$
	\end{algorithmic}
\end{algorithm}
\begin{algorithm}[tb]
	\caption{The CE local graph transformation procedure. It transforms gene subgraph $G$ with parent node $pnode$ to generate child node $cnode$ via CE program $ps$.}
	\label{alg:ceprogram}
	\begin{algorithmic}[1]
		\Procedure{Transform}{$G, ps, pnode, cnode$}
		\If {$ps=${\it\qq {SEQ}}}
		\State $succ\gets list(G.successors(pnode))$		
		\For {$node$ \textbf{in} $succ$}
		\State $G.addEdge(cnode, node)$
		\State $G.removeEdge(pnode, node)$
		\EndFor
		\State $G.addEdge(pnode, cnode)$	
		\EndIf
		%SEQ illustration
		\begin{tikzpicture}[remember picture, overlay, line width=0.5pt, yscale=.65pt]
		\tikzstyle{io}=[draw, rounded corners, text width=0.9cm, text centered, text height=5pt, fill=gray!30]
		\tikzstyle{pc}=[draw, text width=1cm, text centered, text height=5pt]
		%\draw[style=help lines, xshift=5cm, yshift=0.7cm] (0,0) grid(1, 3);
		\node[io, xshift=5cm, yshift=0.5cm] (in) at (1,3){$input$};
		\node[pc, xshift=5cm, yshift=0.5cm] (p) at (1,2){parent};
		\node[pc, xshift=5cm, yshift=0.5cm] (c) at (1,1){child};
		\node[io, xshift=5cm, yshift=0.5cm] (out) at (1,0){$output$};
		\draw[->] (in) edge (p)
		(p) edge (c)
		(c) edge (out);
		\end{tikzpicture}
		%\Statex		
		\If {$ps=${\it\qq {CPI}}}
		\State $pred = list(G.predecessors(pnode))$
		\State $succ = list(G.successors(pnode))$
		\For {$node$ \textbf{in} $pred$}
		\State $G.addEdge(node, cnode)$
		\EndFor
		\For {$node$ \textbf{in} $succ$}
		\State $G.addEdge(cnode, node)$
		\State $G.removeEdge(pnode, node)$
		\EndFor
		\State $G.addEdge(pnode, cnode)$
		\EndIf
		%CPI illustration
		\begin{tikzpicture}[remember picture, overlay, line width=0.5pt, yscale=.65pt]
		\tikzstyle{io}=[draw, rounded corners, text width=0.9cm, text centered, text height=5pt, fill=gray!30]
		\tikzstyle{pc}=[draw, text width=1cm, text centered, text height=5pt]
		%\draw[style=help lines, xshift=5cm, yshift=2.1cm] (0,0) grid(1, 3);
		\node[io, xshift=5cm, yshift=1.4cm] (in) at (1,3){$input$};
		\node[pc, xshift=5cm, yshift=1.4cm] (p) at (1,2){parent};
		\node[pc, xshift=5cm, yshift=1.4cm] (c) at (1,1){child};
		\node[io, xshift=5cm, yshift=1.4cm] (out) at (1,0){$output$};
		\draw[->] (in) edge (p)
		edge[bend right=68] (c)
		(p) edge (c)
		(c) edge (out);
		\end{tikzpicture}
		%\Statex
		\If {$ps=${\it\qq {CPO}}}
		\State $succ = list(G.successors(pnode))$
		\For {$node$ \textbf{in} $succ$}
		\State $G.addEdge(cnode, node)$
		\EndFor
		\State $G.addEdge(pnode, cnode)$
		\EndIf
		%CPO illustration
		\begin{tikzpicture}[remember picture, overlay, line width=0.5pt, yscale=.65pt]
		\tikzstyle{io}=[draw, rounded corners, text width=0.9cm, text centered, text height=5pt, fill=gray!30]
		\tikzstyle{pc}=[draw, text width=1cm, text centered, text height=5pt]
		%\draw[style=help lines, xshift=5cm, yshift=0.5cm] (0,0) grid(1, 3);
		\node[io, xshift=5cm, yshift=0.3cm] (in) at (1,3){$input$};
		\node[pc, xshift=5cm, yshift=0.3cm] (p) at (1,2){parent};
		\node[pc, xshift=5cm, yshift=0.3cm] (c) at (1,1){child};
		\node[io, xshift=5cm, yshift=0.3cm] (out) at (1,0){$output$};
		\draw[->] (in) edge (p)
		(p) edge (c)
		edge[bend right=68] (out)
		(c) edge (out);
		\end{tikzpicture}
		\EndProcedure
	\end{algorithmic}
\end{algorithm}
\subsubsection{Evolutionary Process}
We adopt the evolutionary process in GEP and refer Ferreira \cite{ferreira_2006:gep} for details of the steps presented here. 
\setlength{\hangingindent}{1.5em}
\begin{compacthang}
	\item Step 1: {\it Initialization}\textemdash randomly generate a population of SLGE chromosomes in a uniform distribution manner. 
	\item Step 2: {\it Mapping}\textemdash apply the mapping function $\mathcal{F}_g$ to translate individual chromosomes into cells. 
	\item Step 3: {\it Fitness}\textemdash build candidate CNN architectures with each cell, and train each via back-propagation and evaluate its fitness using the fitness function $\mathcal{F}_p$.
	\item Step 4: {\it Selection}\textemdash select individuals to form next generation population via roulette-wheel strategy with elitism.
	\item Step 5: {\it Mutation}\textemdash randomly mutate all elements in a chromosome. The structural rule must be preserved (e.g. convolution element cannot be assigned to a gene head).	
	\item Step 6: {\it Inversion}\textemdash randomly invert some sequence of elements in a gene head of individual chromosomes.
	\item Step 7: {\it Transposition}\textemdash randomly replace some sequence of elements with  consecutive elements in the same chromosome. The structural rule must be preserved.
	\item Step 8: {\it Recombination}\textemdash crossover gene elements of two chromosomes using two-­point. Since SLGE chromosomes are fixed-length strings similar to the ones in GEP, performing two-point crossover always yields a valid chromosome.
	\item Step 9: Go to Step 2 if max generation not met; else return individual with the highest fitness as best discovered cell.	
\end{compacthang}
\section{Experiments}
The preliminary experiments aimed at verifying the effectiveness of the proposed method to discover cells for CNN architectures that perform well on image classification tasks. We ran eight experiments, two each of four different configurations of chromosome (Table \ref{tab:result1}) and a random search as baseline. The search was conducted on CIFAR-10~\cite{krizhevsky_website_2009} dataset, and the best discovered architecture  was transferred to CIFAR-100~\cite{krizhevsky_website_2009}. The experiments were performed on one 11GB GPU GeForce GTX 1080 Ti machine for 20 days. We implemented all the architectures with PyTorch\footnote{https://pytorch.org/} and trained using fastai\footnote{https://docs.fast.ai/} library. The codes are available at https://github.com/cliffbb/geppy\_nn. %\ref{https://github.com/cliffbb/geppy\_nn}{https://github.com/cliffbb/geppy\_nn}.
\subsection{Evolutionary Settings}
We used a small population of 20 individuals since GEP is capable of solving relatively complex problems with small population \cite{Ferreira2002}, and generation of 20 to reduce the computation cost of the search process.  Other evolutionary parameters settings used for each run are summarized in Table \ref{tab:evo_params}, and these are defaults settings in GEP.
\begin{table}[t]
	\centering
	\begin{tabular}{lc}
		\toprule
		Parameter & Value \\
		\toprule
		%Number of generations & 20 \\
		%Population size & 20 \\
		%Length  of a gene head & 2, 3  \\
		%Number of genes in chromosome & 2, 3  \\
		Number of elites & 1  \\
		Mutation rate & 0.044 \\
		Inversion and Transposition rate & 0.1 \\
		Inversion and Transposition elements length & 2  \\
		%Transposition rate & 0.1  \\
		Two-point/Gene recombination rate & 0.6/0.1 \\
		%Gene recombination rate & 0.1  \\
		\bottomrule
	\end{tabular}
	\caption{Evolutionary parameters for the search process.}
	\label{tab:evo_params}
\end{table}
\begin{table}[t]
	\centering
	\begin{tabular}{NNLL}
		\toprule
		Chromosome & Network & Error & Params \\
		\toprule
		\multirow{2}{*}{Genes=2, Head=2} & $B$=[3, 3, 2] & 3.93 & 3.2M\\ & $B$=[3, 3, 2] & 3.84 & 2.7M\\
		%\midrule
		\multirow{2}{*}{Genes=2, Head=3} & $B$=[3, 3, 1] & \textbf{3.74} & \textbf{2.8M}\\ & $B$=[2, 3, 1] & 3.85 & 2.7M\\
		%\midrule
		\multirow{2}{*}{Genes=3, Head=2} & $B$=[3, 3, 1] & 4.24 & 3.5M\\ & $B$=[3, 3, 1] & 4.03 & 3.4M\\
		%\midrule
		\multirow{2}{*}{Genes=3, Head=3} & $B$=[3, 3, 2] & 4.19 & 3.4M\\ & $B$=[3, 3, 2] & 5.07 & 3.0M\\
		\bottomrule
	\end{tabular}
	\caption{Experimental results on four different configurations of chromosome. We ran evolutionary search on each configuration twice, after search, trained from scratch with $C$=40 and report classification error rate (\%) on CIFAR-10. The boldface is the best result.} 
	\label{tab:result1}
\end{table}
\begin{table*}[t]
	\centering
	\begin{tabular}{lrCCLcl}
		\toprule
		\multicolumn{2}{l}{Model} & CIFAR-10 & CIFAR-100 & Params  & GPU-days & Method \\
		\toprule
		\multicolumn{2}{l}{ResNet-110~\cite{he_2016:drl}}
		& 6.61 & -- & 1.7M & -- & \multirow{5}{*}{Handcrafted} \\
		\multicolumn{2}{l}{ResNet-1001 (pre-activation)~\cite{He2016:IMDRN}} 
		& 4.62 & 22.71 & 10.2M & -- &\\
		\multicolumn{2}{l}{FractalNet~\cite{larsson2017fractalnet}}& 5.22 & 23.30 & 38.6M & -- &\\   
		\multicolumn{2}{l}{DenseNet ($k$=12)~\cite{huang_2017:dcnn}} & 4.10 & 20.20 & 7.0M & -- &\\
		%\multicolumn{2}{l}{DenseNet ($k$=24)~\cite{huang_2017:dcnn}} & 3.74  & 19.25 & 27.2M & -- &\\   
		\multicolumn{2}{l}{DenseNet-BC ($k$=40)~\cite{huang_2017:dcnn}} & 3.46 & 17.18 & 25.6M & -- &\\ 
		\midrule  
		\multicolumn{2}{l}{MetaQNN (top model)~\cite{baker_2017:d-nna-rl}} & 6.92 & 27.14 & 11.2M & 100 & \multirow{6}{*}{Reinforcement} \\
		\multicolumn{2}{l}{NASv2~\cite{zoph_2017:nas-rl}}& 6.01 & -- & 2.5M &  22,400& \\
		\multicolumn{2}{l}{NASv3~\cite{zoph_2017:nas-rl}} & 4.47 & -- & 7.1M & 22,400 &\\
		\multicolumn{2}{l}{Block-QNN-S (more filters)~\cite{zhong_2018:p-nnag}}
		& 3.54 & 18.06 & 39.8M & 96 &\\ 
		\multicolumn{2}{l}{NASNet-A (6 @ 768)~\cite{zoph_2018:lta}}& 3.41 & -- & 3.3M & 2,000 & \\
		\multicolumn{2}{l}{ENAS + cutout~\cite{pham_2018:e-nas-ps}}& 2.89 & -- & 4.6M &  0.5 & \\
		%\multicolumn{2}{l}{NASv3~\cite{zoph_2017:nas-rl}} & 4.47 & -- & 7.1M & 22,400 &\\
		
		\midrule  
		\multicolumn{2}{l}{DARTS (first order) + cutout ~\cite{liu_2019:dart}} & $3.00\pm0.14$   & -- & 3.3M & 1.5 & \multirow{4}{*}{ Gradient-Based} \\
		\multicolumn{2}{l}{DARTS (second order) + cutout~\cite{liu_2019:dart}}& $2.76\pm0.09$ & -- & 3.3M &  4 & \\
		\multicolumn{2}{l}{P-DARTS + cutout~\cite{chen:2019pdarts}}& 2.50 & -- & 3.4M &  0.3 & \\
		\multicolumn{2}{l}{P-DARTS + cutout~\cite{chen:2019pdarts}}& --  & 17.20 & 3.4M &  0.3 & \\
		
		\midrule
		\multicolumn{2}{l}{Genetic CNN~\cite{xie_2017:g-cnn}} & 7.10 & 29.05 & -- & 17 &\multirow{7}{*}{Evolutionary}\\
		\multicolumn{2}{l}{Large-scale Evolution~\cite{real_2017:l-eic}} & 5.40 & -- & 5.4M & 2,750 &\\
		\multicolumn{2}{l}{Large-scale Evolution~\cite{real_2017:l-eic}} & -- & 23.0 & 40.4M & 2,750 &\\
		\multicolumn{2}{l}{Hierarchical Evolution~\cite{liu_2018:hr-eas}} & 3.63 & -- & -- & 300 &\\ 
		%\multicolumn{2}{l}{CGP-CNN~\cite{suganuma_2018:gp-cnna}} & 5.01 & -- & 1.7M &  (2 GPUs) &\\
		%\multicolumn{2}{l}{CGP-CNN~\cite{suganuma_2018:gp-cnna}} & -- & 26.5 & 4.6M & (2 GPUs) &\\ 
		\multicolumn{2}{l}{CA-CNN~\cite{sun_2019a:ca-cnn}} & 4.30 & -- & 2.0M & 27 &\\
		\multicolumn{2}{l}{CA-CNN~\cite{sun_2019a:ca-cnn} } & -- & 20.85 & 5.4M & 36 &\\
		\multicolumn{2}{l}{AmoebaNet-A~\cite{real_2019:re-icas}} & 3.34 & -- & 3.2M & 3,150 &\\
		\midrule
		\multirow{3}{2.5cm}{SLGE Networks (our approach)} 
		& $C$=40, $B$=[3, 3, 1] & 3.74 & 22.95 & 2.8M & 20 & Evolutionary \\
		& $C$=40, $B$=[3, 3, 2] & 3.84 & 24.47 & 2.7M & 20 & Evolutionary \\
		& $C$=40, $B$=[4, 4, 3] & 4.47 & -- & 3.2M & 4 &Random search \\ 
		\bottomrule
	\end{tabular}
	\caption{Comparisons between the proposed method and other architecture search methods in terms of classification error (\%), number of parameters and computing days of GPU ({\it \# of employed GPUs $\times$ \# of days the algorithms	was run}) based on CIFAR-10 and CIFAR-100 datasets.}
	\label{tab:result2}
\end{table*}
\subsection{Training Details}
Each network begins with 3$\times$3 convolution stem with output channel size $C$, followed by three blocks denoted as $B$=$[b_1,b_2,b_3]$\textemdash each block $i$ consist of repeated cell of $b_i$ times with max-pooling layer inserted between them to downsample\textemdash then a classifier. The max-pooling reduces the feature maps of each block by a half. The channels are doubled after each max-pooling operation.

During search, the CIFAR-10 50k training set was split into 40k training and 10k validation subsets with normalization. We used relatively small network with $C$=16 and $B$=[1, 1, 1]. Each candidate network was trained on the training subset and evaluated on the validation subset to determine its fitness value using 1-cycle policy in Smith~\cite{Leslie_2018:onecycle} with Adam optimizer. The learning rate was set to go from 0.004 to 0.1 linearly while the momentum goes from 0.95 to 0.85 linearly in phase one, then in phase two, the learning rates follows cosine annealing from 0.1 to 0, whereas the momentum goes from 0.85 to 0.95 with the same annealing. The weight decay was set to 0.0004, batch size to 128 and training epochs to 25. 

After the search, we set $C$=40 and each evolutionary-discovered cell was repeatedly stacked to build large CNN architectures subject to model size constraint $\mathcal{T}_{params}$$\leq$3.5M. We trained each CNN architecture from scratch on the training and validation subsets for 800 epochs, and reported the classification error on the CIFAR-10 10k test dataset. The learning rate remained as during search for the first 350 epochs, and was reset to go from 0.00012 to 0.003 in phase one and 0.003 to 0 in phase two of 1-cycle policy. We augmented the training subset as in He {\it et al.}~\cite{he_2016:drl}, and the other hyperparameters remained the same as during search.
\subsection{Search on CIFAR-10}
To investigate the effectiveness of SLGE, we conducted evolutionary search on four different configurations of chromosome\textemdash2 genes with head of 2 and 3, and 3 genes with head of 2 and 3. Each was run twice on CIFAR-10 dataset. Chromosome of 2 genes with head of 3 discovered the best architecture which obtains 3.74\% classification error (Table \ref{tab:result1}). The result is compared with other search methods in Table \ref{tab:result2}.
\paragraph{Compared with handcrafted networks} SLGE networks improve the performance of most popular handcrafted ones. SLGE achieves classification error of 0.8\% and 1.4\% lower than ResNet-1001 and FractalNet respectively with few parameters. However, the DenseNet-BC ($k$=40) performs better than SLGE networks using more parameters.
\paragraph{Compared with reinforcement NAS networks} It shows that SLGE achieves the error rate of approximately 3.2\% and 2.3\% improvement to MetaQNN and NASv2 respectively; and competes with Block-QNN-S (used more parameters) and NASNet-A (consumed more GPU resources). With cutout augmentation and 1.8M parameters more, ENAS achieved 0.9\% error rate lower than SLGE using 0.5 GPU days.
\paragraph{Compared with gradient-based NAS networks} Using cutout augmentation method, DARTS and P-DARTS used less GPU days to achieve classification error rate of approximately 0.9\% and 1.2\% lower than SLGE respectively. However, the SLGE networks have fewer parameters than DARTS and P-DARTS.
\paragraph{Compared with evolutionary NAS networks} SLGE performs significantly well than Genetic CNN and Large-scale Evolution, but falls slightly behind the state-of-the-art Hierarchical Evolution  and AmeobaNet-A networks. SLGE consumes 0.06 and 0.006 GPU computing days of that consumed by Hierarchical Evolution  and AmeobaNet-A respectively.
\subsection{Random Search}\enlargethispage{0.5\baselineskip} 
To evaluate the effectiveness of the SLGE representation scheme, a simple random search was performed as the baseline. We randomly generated a population of ten individual networks from the top chromosome (Table \ref{tab:result1}); trained all from scratch with no evolutionary modifications on CIFAR-10 using the same training settings with 500 epochs. The network with the lowest classification error is reported in Table \ref{tab:result2}. We achieve the mean classification error of 5.85\%  with standard deviation 1.77\% over the ten individual networks. This is a considerable competitive result which demonstrates the effectiveness of our proposed SLGE scheme with regular convolution search units.
\subsection{Evaluation on CIFAR-100}
The top architecture learned on CIFAR-10 task was experimented on CIFAR-100 to evaluate the transferability of the evolutionary-discovered cell. The CIFAR-100 has similar features as CIFAR-10, but with 100 classes which makes it rather challenging classification task. We simply transferred the best architecture from CIFAR-10 (Table \ref{tab:result1}) but trained from scratch with a classifier head of 100 classes, all training settings remained the same. And we achieve the classification error of 4.1\%  improvement to MetaQNN and 6.1\% to Genetic CNN, and compete with other networks (Table \ref{tab:result2}).
\subsection{Discussion}\label{sec:discussion}
We make a few observations from the preliminary experimental results. The networks developed with chromosomes of 2 genes performed better than the ones with 3 genes, something we least expected.  And the evolutionary-discovered cell (Fig. \ref{fig:cell}) of the top network has topological structure similar to DenseNet with a well-formed asymmetric convolution. This underscore the effectiveness of the proposed method and further experiments will be conducted in future to ascertain  its robustness.	In general the preliminary results on CIFAR-10 and CIFAR-100 classification tasks are very promising, although no evolutionary parameter was tuned. Thus, the natural future direction is to tune the evolutionary parameters, and extend the task to the most challenging ImageNet classification task to verify the generality of the proposed method. 
\section{Conclusion}
We have presented an effective evolutionary method which discovers high-performing cells as building blocks for CNN architectures based on a representation scheme which embeds elements of CE in chromosomes of GEP. We show that our SLGE representation scheme coupled with regular convolution units can achieve significant results even with a simple random search strategy. Our top network obtains a highly competitive performance with the state-of-the-art networks on both CIFAR-10 and CIFAR-100 datasets. In addition to the observations made in Section \ref{sec:discussion}, we will continue improving the proposed method and extend it to other visual tasks such as semantic segmentation and object detection.

\bibliographystyle{IEEEtran}
\bibliography{ssci_ref}

\end{document}